\documentclass{article}

\usepackage{PRIMEarxiv}

\usepackage[utf8]{inputenc} 
\usepackage[T1]{fontenc}    
\usepackage{hyperref}       
\usepackage{url}            
\usepackage{booktabs}       
\usepackage{amsfonts}       
\usepackage{nicefrac}       
\usepackage{microtype}      
\usepackage{lipsum}
\usepackage{fancyhdr}       
\usepackage{graphicx}       
\usepackage{amsmath}
\usepackage{multirow}
\usepackage{verbatim}
\usepackage{xcolor}
\usepackage{soul}
\DeclareMathOperator*{\argmax}{arg\,max}

\graphicspath{{media/}}     

\title{A molecular generative model with genetic algorithm and tree search for cancer samples}

\author{
  Sejin Park \\
  School of Electrical Engineering and Computer Science \\
  Gwangju Institute of Science and Technology \\
  Gwangju 61005, South Korea.\\
  \texttt{sejin8544l@gm.gist.ac.kr} \\
   \And
  Hyunju Lee\,$^{*}$ \\
  School of Electrical Engineering and Computer Science \\
  Gwangju Institute of Science and Technology \\
  Gwangju 61005, South Korea.\\
  \texttt{hyunjulee@gist.ac.kr} \\
}

\begin{document}
\maketitle

\begin{abstract}
Personalized medicine is expected to maximize the intended drug effects and minimize side effects by treating patients based on their genetic profiles. Thus, it is important to generate drugs based on the genetic profiles of diseases, especially in anticancer drug discovery. However, this is challenging because the vast chemical space and variations in cancer properties require a huge time resource to search for proper molecules. Therefore, an efficient and fast search method considering genetic profiles is required for de novo molecular design of anticancer drugs. Here, we propose a faster molecular generative model with genetic algorithm and tree search for cancer samples (FasterGTS). FasterGTS is constructed with a genetic algorithm and a Monte Carlo tree search with three deep neural networks: supervised learning, self-trained, and value networks, and it generates anticancer molecules based on the genetic profiles of a cancer sample. When compared to other methods, FasterGTS generated cancer sample-specific molecules with general chemical properties required for cancer drugs within the limited numbers of samplings. We expect that FasterGTS contributes to the anticancer drug generation
\end{abstract}

\section{Introduction}
Traditional drug discovery and development are very time-consuming (approximately 12 years) and expensive (2.7 billion USD) because of the high cost of clinical trial failures \cite{smietana2016trends, mullard20182017}. One of the main reasons for this is the enormous chemical space of drug development, where the number of potential molecules is almost 10${^{60}}$ \cite{kirkpatrick2004chemical, dobson2004chemical, hert2009quantifying}. Nevertheless, for the past two decades, high throughput screening, with hit rates of 0.01–0.14\%, has been the main method of drug discovery \cite{bender2008aspects, zhu2013hit, goodwin2020statistical}. Because conventional methodologies are expensive and less effective, computational methods based on artificial intelligence techniques and machine learning methods have been increasingly used to improve drug development efficiency \cite{sanchez2018inverse, walters2020assessing, schneider2020rethinking}.

Although deep learning models are very powerful, some problems, such as overfitting, can occur if the training datasets are not large enough. Overfitting prevents models from achieving similar performance in test data compared to training data. In addition, it is difficult to obtain unique molecules from generative models trained on a small dataset.

As labeled datasets are insufficient for drug discovery, reinforcement learning (RL) is commonly used to optimize the target property instead of supervised learning (SL) \cite{olivecrona2017molecular, popova2018deep, zhou2019optimization, staahl2019deep, born2020paccmann}. \cite{born2020paccmann} used two encoder-decoder architectures for compounds and gene expression profiles; gene expression latent space was added to the molecule latent space to generate anticancer drugs, and the model was trained by RL. \cite{mendez2020novo} proposed a generative adversarial network-based model \cite{goodfellow2014generative}, where the encoder-decoder architecture was used with gene expression signatures. 

Although many methods used RL to optimize target property, reward-based RL highly provokes overfitting \cite{renz2020failure}. CogMol \cite{chenthamarakshan2020cogmol} was designed to generate drugs for COVID-19, and it used the controlled latent attribute space sampling method \cite{das2020accelerating} instead of RL.  Also, Monte Carlo tree search (MCTS) \cite{yang2017chemts, srinivasan2020artificial, kajita2020autonomous} and a genetic algorithm (GA) \cite{jensen2019graph, ahn2020guiding, nigam2021janus} were employed to create a molecular generative model. Especially, \cite{ahn2020guiding} developed  a generative model using GA and primary queues. Their model could generate highly-rewarding molecules but requires extensive samplings to achieve high performance.

The success rates of anticancer drug investigations in clinical development are three times lower than those of cardiovascular disease-related drugs \cite{kola2004can, roberts2004trends}. One of the main reasons is that anticancer drugs tend to target not only cancer-related genes in tumor cells but also other essential genes in normal cells \cite{kamb2007cancer}. In general, a low IC${_{50}}$ or GI${_{50}}$ value means that an anticancer drug reacts well to a given cell line; thus, several studies \cite{born2020paccmann, joo2020generative} used these values of cell lines to evaluate candidate cancer drugs. However, this approach has limitations for generating molecules for precision medicine. First, some molecules with low IC${_{50}}$ values may cause variations in cellular behavior for the entire cells instead of target cancer cells. Second, generated molecules highly optimized on IC${_{50}}$ values are more likely to be overfitted \cite{renz2020failure} and might not have proper drug properties. In contrast, the performance might not be satisfactory when aiming to avoid overfitting. Thus, to evaluate the generated anticancer molecules, the relative IC${_{50}}$ values of a given cancer sample against  IC${_{50}}$ values of other samples and drug properties should be considered. 

In this study, to generate anticancer molecules based on the genetic profiles of a cancer sample, we propose a FasterGTS approach constructed with GA and MCTS. Here, supervised learning and self-trained networks are used to generate molecules for cancer samples, and a value network is used to predict their IC${_{50}}$ values based on multi-omics data of the cancer sample. In this process, the combination of GA and MCTS complements each other. To be specific, GA requires many samplings to generate proper molecules, and it is hard for MCTS to generate different molecules from those already generated. However, the combination is helpful to generate various molecules satisfying targeting properties in a small number of samplings. Furthermore, in MCTS, we evaluate uncompleted molecules using rollout during tree expansion, and the evaluation becomes more reliable with iterations. This way of  using MCTS was inspired by AlphaGo \cite{silver2016mastering}, which maximizes the performance by recording each step on the tree so that  the entire system can use the history to explore experimental ways for developing unique results.

The contributions of FasterGTS are as follows. First, FasterGTS can generate anticancer drugs for a target sample based on multi-omics data. The cancer sample-specific drug has a lower IC${_{50}}$ value for the target sample than other samples. Second, we improve the model performance within a restricted condition by using GA and a self-trained network with MCTS. Here, the good performance means that the molecules have a high reward score and satisfy the general drug property requirements that are not included in the reward term.

\begin{figure*} [t]
\includegraphics[width=\textwidth,height=4cm]{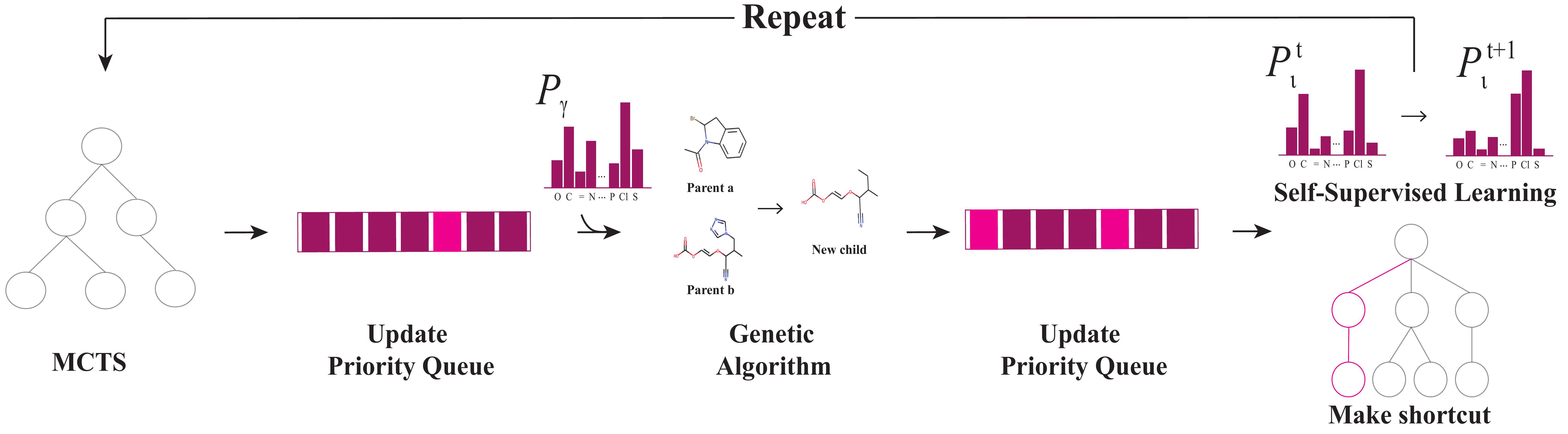} 
  \caption{The overall workflow of FasterGTS for generating cancer sample-specific drugs. } \label{fig1}
\end{figure*} 
\section{Materials and Methods}
\subsection{Datasets}
The ChEMBL \cite{bento2014chembl} database has approximately 1.5 million SMILES strings, not only limited to cancer drugs. Among them, we used samples less than 100 characters long. We used 223 cancer drugs of the Genomics of Drug Sensitivity in Cancer (GDSC) database \cite{yang2012genomics} and 561 cell lines in the Cancer Cell Line Encyclopedia (CCLE) \cite{barretina2012cancer}, where the numbers of genes in gene expression, mutation, and methylation datasets were 697, 710, and 808, respectively. The IC${_{50}}$ values of drugs in GDSC were used as the labeled data \cite{iorio2016landscape}, and the pairs of drugs and cell lines are 107,446.   

\subsection{Overview of FasterGTS}
The objective of this study is to generate cancer cell-specific drugs based on the genetic characteristics of a given cancer cell. The proposed FasterGTS employs MCTS and GA for generating cancer-cell specific drugs. In MCTS, a general drug (GD) policy network ${p_\gamma}$, a cancer sample-specific drugs (CD) policy network ${p_\chi}$, and a value network  ${v_{\theta}}$ are main components to generate or evaluate molecules. The GD policy network ${p_\gamma}$ is trained by SMILES strings in ChEMBL, and thus most of the generated drugs by ${p_\gamma}$ are not specific to the given cancer sample. To evaluate that generated drugs are effective for the given cancer sample, a value network ${v_{\theta}}$ is constructed.  The value network ${v_{\theta}}$  is trained using cancer drugs from GDSC and cell lines from CCLE, and genetic characteristics of the cancer cell such as gene expressions and methylations are used as important features to predict the efficacy (IC$_{50}$ values) of the generated drugs. In addition, another generative model called a CD policy network ${p_\chi}$ is constructed by self-training, where the GD policy network ${p_\gamma}$ is fine-tuned with molecules with high reward values for the given cancer sample. 

The workflow of the proposed approach is shown in Fig. 1. First, FasterGTS searches proper molecules with MCTS and updates a priority queue with molecules having higher rewards than those stored in the queue. Second, molecules randomly selected from the queue and those generated by the GD policy network are used as parents for GA, and the queue is updated if there are new molecules having higher rewards than stored ones. Finally, FasterGTS trains the CD policy network ${p_\chi}$ with molecules randomly selected from the queue and makes shortcuts in MCTS with molecules, which are newly generated by GA and stored in the priority queue. This process is repeated.

In the following subsections, we first introduce the GD policy network  ${p_\gamma}$ and the value network  ${v_{\theta}}$, then explain MCTS, followed by a priority queue, the CD policy network, and the GA.

\subsection{GD policy network}
\label{GD}
The GD policy network ${p_{\gamma}}$ generates drug-like molecules, which have general properties of drugs. We use two different generative models for ${p_{\gamma}}$, stack-augmented recurrent neural networks (RNNs) \cite{joulin2015inferring} and Generative Pre-trained Transformer (GPT) \cite{radford2018improving}. GPT is the decoder of the transformer \cite{vaswani2017attention}, which consists of fully connected networks and an attention-based encoder-decoder model. 

Stack-augmented RNNs store and process useful information from the SMILES string to learn complex rules of the drug-like molecule SMILES string better than simple RNN-based models \cite{segler2018generating, gupta2018generative}. \cite{popova2018deep} used stack-augmented RNNs for a drug-like molecule generative model, and we used the same architecture as theirs in our experiment. 

GPT is an advanced model commonly used for natural language processing tasks because attention-based models can effectively utilize the information of previous sentences and reduce the risk of the vanish gradient problem. \cite{bagal2021liggpt} and \cite{kim2021generative} utilized GPT for a molecular generative model. A conditional token and latent space are used to optimize a target property for \cite{bagal2021liggpt} and \cite{kim2021generative}, respectively. We referred to the code of \cite{bagal2021liggpt}, but our GPT is a vanilla GPT without the conditional vector.

When RNNs and GTP are trained, stochastic gradient ascent is used to maximize the likelihood of the next action ${a}$ producing a drug-like molecule in state ${s}$. Because each action, a character, can be seen as one of the classes, the cost function is a cross-entropy loss function, and the cost function ${J_\gamma(d)}$ with the network parameters ${\gamma}$ is minimized. 
\begin{equation} \label{SSL_eq1}
\Delta \gamma \propto \frac{\partial \log{p_\gamma(a|s)}}{\partial \gamma}
\end{equation}
\begin{equation} \label{SSL_eq2}
J_\gamma(s_T) = -\sum_{t=1}^{T}\sum_{i=1}^{|A|}c_i\log{(p_\gamma(a_i|s_{t-1}))},
\end{equation}
where ${s_T}$, ${T}$, and ${c}$ are the SMILES string of a terminal state, the length of the SMILES string, and a true character label, respectively.

\begin{figure*} [t]
\includegraphics[width=\textwidth,height=4.5cm]{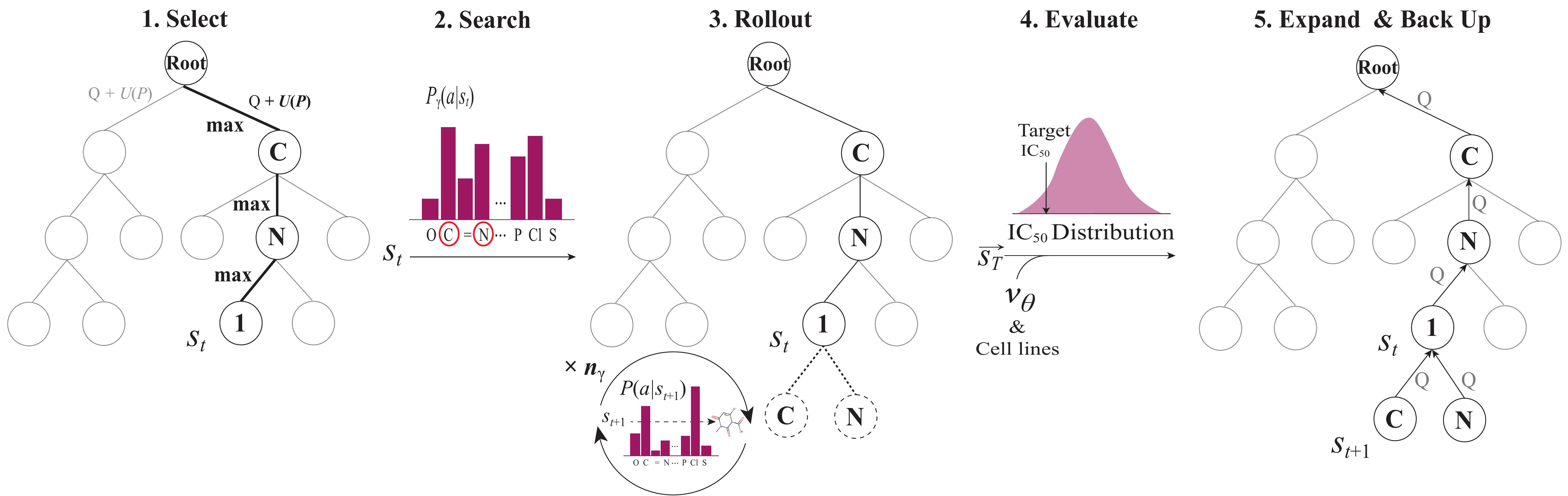} 
  \caption{The workflow of MCTS in FasterGTS at ${s_{t}}$. Herein, the numbers are their sequence, and each node shows their action, i.e., the last character of their state. First, FasterGTS selects the best node among brother nodes based on the tree policy (Equation (\ref{tree_policy})). FasterGTS searches possible next actions, i.e., characters, based on ${p_\gamma}$ given ${s_{t}}$. For evaluation, FasterGTS performs rollout of each action ${n_r}$ times to generate terminal state ${s_T}$, completed SMILES, from non-terminal state ${s_{t+1}}$, uncompleted SMILES, using ${p_\chi}$ and ${p_\gamma}$ given ${s_{t+1}}$. To be specific, most molecules are generated by ${p_\chi}$ and the others by ${p_\gamma}$ to maintain diversity. Next, FasterGTS evaluates ${s_{t+1}}$ with the terminal states vector in terms of the reward (Equation (\ref{reward_policy})). During the evaluation, adversary cell lines are used to build IC${_{50}}$ distribution, using the value network, ${v_{\theta}}$, and the target sample $z$-score is calculated from the distribution. Finally, FasterGTS expands new actions as new nodes and backs up the new information until the root node.} \label{fig2}
\end{figure*} 

\subsection{Value network}
In FasterGTS, DeepCDR \cite{liu2020deepcdr} is used for the value network ${v_{\theta}}$. For a cell line $k$, the inputs are multi-omics data and SMILES, and the output is a predicted IC${_{50}}$ value.  
\begin{equation}
y_k = v_{\theta}(d,\mathrm{c}_k),
\end{equation}
where ${y_k}$, ${d}$, and ${\mathrm{c}_k}$ are the predicted IC${_{50}}$ value, SMILES string, and a cell line vector, respectively.
\begin{equation}
\mathrm{c}_k = [\mathrm{g}_k; \mathrm{mu}_k; \mathrm{me}_k],
\end{equation}
where ${\mathrm{g}_k}$, ${\mathrm{mu}_k}$, and ${\mathrm{me}_k}$ are the gene expression, mutation, and methylation vectors, respectively.
The value network ${v_{\theta}}$ consists of a uniform graph convolutional network and multiple subnetworks \cite{liu2020deepcdr}. The DeepChem library \cite{Ramsundar-et-al-2019} is used to obtain the node features and graph of a given SMILES string.

\subsection{MCTS}
A molecular space specific for a cancer cell might be narrow. Thus, if the GD policy space ${p_\gamma}$ is used to generate candidate drugs, a large number of samplings are required to obtain molecules whose predicted IC$_{50}$ values are low enough to a target cancer cell. To alleviate that, we use MCTS, consisting of the GD policy network ${p_\gamma}$, the CD policy network ${p_\chi}$ (details are in subsection \ref{CD}), and the value network ${v_{\theta}}$. The ${p_\gamma}$ has a large search space for various actions, but the ${p_\chi}$ has a small search space for a target sample. In general, a good searching strategy needs to seek not only promising actions but also challenging ones and an effective evaluation needs to find the most suitable action for anticancer drugs. Considering these properties, the GD policy network is used to prevent overfitting and maintain diversity, and the CD policy network is used to find cancer cell specific molecules. In the evaluation, the value network ${v_{\theta}}$ is used as a critic.

\subsubsection{Tree Search}
MCTS is constructed with a node ${(s,a)}$, where the state ${s}$ is a current molecular string and an action ${a}$ is a chosen character and the last character of the state. In our model, the states with uncompleted and completed SMILES strings are called non-terminal and terminal states, respectively. Therefore, we can define sets of possible actions, states, and terminal states as ${A}$, ${S}$, and ${S^T=}$\{${s_T\in S}$\}, respectively. The MCTS mainly operates in three steps: selection, searching, and evaluation. Selection is to choose the best node at each depth (based on Eq. (\ref{tree_policy})). Searching explores the new next nodes of a non-terminal state based on prior probability, which is from the distribution of the generative model at the current state. Evaluation gives a quality score to new nodes, using molecules generated from rollout, which creates terminal states from a non-terminal state of the node to be assessed. 

A node ${(s,a)}$ on MCTS has a set of statistics, ${\{ P, N, S, N_v, N_w, Q\}}$, which are prior probability, the visit count, the sum of rewards, the total valid molecule count, the total winning count, and the action value, respectively. The prior probability is a probability to be the current state from the previous state, the total valid molecule count means the sum of valid molecules generated from rollout (in Eq. (\ref{total_valid_count})), and the total winning count is the number of valid molecules with a reward value greater than one (in Eq. (\ref{total_winning_count})). The action value is the score to represent how proper the node is to be cancer sample-specific cancer drugs (in Eq. (\ref{action_value})).

\subsubsection{Evaluation and Backup}
From the value network ${v_{\theta}}$, we can only obtain the reward of completed SMILES on the terminal state ${s_T}$, but not for the non-terminal state ${s_t}$, ${0<t<T}$. To obtain an immediate reward from ${s_t}$, rollout is used, which generates a completed SMILES string from an uncompleted one of ${s_t}$.  For rollout, the network ${p_{\chi}}$ is used to generate anticancer molecules on a given cell line, and ${p_{\gamma}}$ is also used for diversity. Although the results of the rollout on the initial shallow nodes ${(s_t,a)}$, ${0<t \ll T}$, do not precisely evaluate ${s_t}$, we can obtain more reliable information from the shallow nodes by updating the search tree. A rollout function ${\rho_\chi}$, which derives terminal states from a non-terminal state, is defined as follows, and  the terminal states vector  ${\vec{s_{T}}}$ and  the number of valid molecules $n_v$ are obtained.
\begin{equation}
\begin{split}
[\vec{s_T}, n_v] = \rho_\chi((s_t,a),n_r),\\ 
\vec{s_T} = [s_{T_1}, s_{T_2} ,..., s_{T_{n_{v}}}]
\end{split}
\end{equation}
where ${n_r}$ is the number of rollout and $n_r \geq n_v$. The completed molecules from the rollout are not always in a valid form. To check whether a generated SMILES string is in a valid form, we used the Chem.MolFromSmiles function in RDKit \cite{Landrum2016RDKit2016_09_4}, which returns the Chem.rdchem.Mol object of the input SMILES string if the input is valid.

To generate cancer cell-specific cancer drugs, we focus on not only a low IC${_{50}}$ value but also on how low the IC${_{50}}$ value of the target sample is compared to those of the other cell samples. The $Z$-score of the IC${_{50}}$ values is a good barometer of how low IC${_{50}}$ value of a generated molecule is, compared to other samples. To build an IC${_{50}}$ distribution, we define the genetic profile of the target cell line vector and those of the adversary cell line matrix as ${\mathrm{c}_t}$ and ${C_a = [c_{a}^1, c_{a}^2,..., c_{a}^{n_a}]}$, respectively. Then, ${y_t}$ and ${y_z}$ that are an IC${_{50}}$ value of a target cell line $t$   and a $Z$-score of ${y_t}$ on other adversary cell lines, respectively, are obtained as follows.
\begin{gather}
\mathrm{y}_a = a_{\theta}(s_{T_k},C_a)\\\nonumber
 =[v_{\theta}(s_{T_k},c_{a}^1), v_{\theta}(s_{T_k},c_{a}^2), ... ,v_{\theta}(s_{T_k},c_{a}^{n_a})]
\end{gather} 
\begin{equation}
\begin{split}
y_t = v_\theta(s_{T_k},\mathrm{c}_t)\\
y_z = \zeta(y_t,\mathrm{y}_a),
\end{split}
\end{equation}
where ${a_\theta, \mathrm{y}_a}$, and ${ \zeta}$ are an adversary reward network, an adversary IC${_{50}}$ value vector, and a $Z$-score function, respectively. Given $s_{T_k}$, ${\mathrm{c}_t}$, and ${C_a}$, the reward function ${r_{\theta}}$ of FasterGTS is defined as follows:
\begin{equation} \label{reward_fun}
R(y_t,y_z)=exp(\alpha(-\Theta_z+y_z))+\beta \times log(-y_t+\Theta_t+1)
\end{equation}

\begin{equation} \label{reward_policy}
r_{\theta}(s_{T_k},\mathrm{c}_t,C_a) = \left\{
     \begin{array}{@{}l@{\thinspace}l}
        R(y_t,y_z) &: y_t \le \Theta_{t}, y_z \le \Theta_{z}\\
        1 &: else,
     \end{array}
   \right.
\end{equation}
where ${\alpha}$ and ${\beta}$ are the constants that balance the IC${_{50}}$ and $Z$-score, respectively, and ${\Theta_{t}}$ and ${\Theta_{z}}$ are the thresholds of the IC${_{50}}$ value and $Z$-score for the reward, respectively. 

The node statistics ${\{S, N_v, N_w, Q\}}$ are updated from the rollout as follows:
\begin{equation}
N_v(s_t, a) = N_v(s_t, a) + n_v
\label{total_valid_count}
\end{equation}
\begin{equation}
S(s_t,a) = S(s_t,a) + \sum_{i=1}^{n_v} r_{\theta}(s_{T_i},\mathrm{c}_t,C_a)
\end{equation}
\begin{equation} \label{indicator_function}
I(x) = \left\{
     \begin{array}{@{}l@{\thinspace}l}
        1 &: x > 1\\
        0 &: else
     \end{array}
   \right.
\end{equation}
\begin{equation}
n_w = \sum_{i=1}^{n_v} I(r_{\theta}(s_{T_{i}},\mathrm{c}_t,C_a))
\end{equation}
\begin{equation} \label{total_winning_count}
N_w(s_t, a) = N_w(s_t, a) + n_w
\end{equation}
\begin{equation} \label{action_value}
Q(s_t,a) = \frac{S(s_t,a)}{1+N_v(s_t,a)},
\end{equation}
where the initial values of $S, N_v, N_w$, and ${Q}$ are zero. In this way, FasterGTS can give the reward to non-terminal nodes. 

Next, in the backup, the parent nodes obtain the properties (${S, n_v}$, and ${n_w}$) of child nodes and update their values using Eqs. (\ref{total_valid_count}, \ref{total_winning_count}, \ref{action_value}). By updating the properties, the search tree can appropriately evaluate future rewards. If all simulation results are invalid, a penalty is applied and updated, where the set is ${ \{ S = -1, n_v = 0, n_w = 0\}}$.

\subsubsection{Selection, searching, and expansion}
Selecting the next action follows the below tree policy, a variant of the PUCT \cite{rosin2011multi}, to choose the best node.
\begin{equation} \label{tree_policy}
    a_t = \argmax_{a\in A} \left(Q(s_t, a)+U(P) \right)
\end{equation}
\begin{equation} \label{bonus_term}
    U(P) = cP(s_t, a)\frac{\sqrt{\sum_{b} N(s_{t-1}, b)}}{1+N(s_t, a)},
\end{equation}
where ${c}$ and ${\sum_{b} N(s_{t-1},b)}$ are the exploration constant and parent visit count, respectively, and the visit count ${N(s_t, a)}$ is incremented by one when the node is selected based on Eq. (\ref{tree_policy}). In  Eq. (\ref{tree_policy}), the first term ${Q(s_t, a)}$ is unlikely to have high scores for less-visited nodes. In contrast, in the second term ${U(P)}$, the rarely visited node's value increases, while the value of the repeatedly visited nodes decays. Therefore, ${p_{\gamma}(a|s)}$ is used as ${P(s,a)}$ because it has a vast search space with the general properties of molecules. If ${P(s,a)}$ is ${p_{\chi}(a|s)}$, ${P(s,a)}$ would focus on the node with a high action value and not guarantee the adventurous paths, from which unexpected and novel results come. This case likely achieves lower IC${_{50}}$ values than when ${P(s,a)}$ is ${p_{\gamma}(a|s)}$; however, our goal is to generate cancer-specific cancer drugs, which have not only low IC${_{50}}$ values but also low ${Z}$-scores.

In the searching, for leaf nodes, we sample new actions ${n}$ times with the probability distribution ${p_{\gamma}}$, and expand the actions as new nodes. Expanding only leaf nodes can reduce the probability of the searching promising actions; thus, the nonleaf nodes with other children nodes are also expanded with a fixed probability.

\subsection{Priority queue, CD policy network, and genetic algorithm}
\label{CD}
Even though MCTS is beneficial to effectively search for proper molecules, it is hard to generate quite different molecules from those already generated. Thus, we utilize GA to examine various molecules. The main processes of GA are crossover and mutation, and molecular crossover and mutation of \cite{jensen2019graph} are used in FasterGTS. In GA, a half of parent molecules are selected from the priority queue and a half are from the GD policy network ${p_\gamma}$. Because ${p_\gamma}$ has a large search space, it is helpful to make various molecules. Molecules generated by GA are stored in the priority queue if they have higher rewards than those in the queue. Since GA generates molecules regardless of the tree policy, MCTS does not have a path related to the molecules from GA. To enhance various searches in the tree, we make paths of the molecules from GA without MCTS steps, and the path is called a shortcut.

Finally, the CD policy network ${p_\chi}$ is self-trained with molecules in the priority queue. ${p_\chi}$ has the same architecture as the GD policy network ${p_{\gamma}}$, such as stack-augmented RNN or GPT, and the initial weights of ${p_\chi}$ are same as those of ${p_{\gamma}}$. However, it is trained using molecules randomly selected from the priority queue to generate new molecules.

\subsection{Comparison methods}

\begin{enumerate}
   \item We compared two versions of FasterGTS depending on the generative model. If the stack-augmented RNN is used, the model is called FasterGTS-RNNs, and if GPT is used, it is called FasterGTS-GPT.
    \item An RL-based approach by \cite{popova2018deep} was compared, where the stack-augmented RNN-based generative model was used. This RNN-based generative model has the same architecture as ${p_{\gamma}}$ and ${p_{\chi}}$. However, \cite{popova2018deep} did not intend to generate cancer-sample specific cancer drugs. Thus, to apply their method to the objective in this study, we trained the generative model twice sequentially: by RL and single-target RL (SRL). RL is the same method used in \cite{popova2018deep}, and SRL uses the same reward function and policy as Eqs. (\ref{reward_fun}) and (\ref{reward_policy}) for a target sample. Thus, SRL is similar to FasterGTS-RNNs, except that it does not use the GA and MCTS. We call this stack-augmented RNNs based generative model \textbf{RNNs}. 

	\item We tested a GPT-based generative model, which has the same architecture as ${p_{\gamma}}$ of FasterGTS-GPT. The generative model was also trained by RL and SRL like RNNs. We name the model \textbf{GPT}. 
	
	\item \textbf{ChemTS} \cite{yang2017chemts} consists of MCTS and an RNNs-based generative model \cite{segler2018generating}, and its goal is to optimize the penalized logarithm of the octanol-water partition coefficient scores \cite{gomez2018automatic}. For a comparison in this study, we used the same generative model ${p_{\gamma}}$, the value function ${v_\theta}$, the reward function, and the policy as those in FasterGTS.
	
	\item We tested a model, which consists of our MCTS and RNNs-based generative model ${p_{\gamma}}$. This model is named \textbf{MCTS-RNNs}. It has a similar environment as ChemTS except for the search tree.  This model allows comparison between MCTS proposed in this study and MCTS used in ChemTS.
	
	\item \textbf{GEGL} \cite{ahn2020guiding} uses a trainable generative model, two priority queues, and GA. Their goal is to optimize properties of the GuacaMol benchmark \cite{brown2019guacamol}. In \cite{ahn2020guiding}, the generative model is based on LSTM and trained on molecular dataset ZINC \cite{irwin2012zinc}. As the performance of a model depends on a molecular dataset used for training a generative model,  we used the same generative model in \cite{ahn2020guiding}, but it was trained using the molecular dataset ChEMBL \cite{bento2014chembl} used in this study.  Because GEGL does not use MCTS, this comparison can show the contribution of the combination of GA and MCTS in FasterGTS.
	
	\item \textbf{JANUS-C} \cite{nigam2021janus} was compared, which uses SELFIES \cite{krenn2020self}, a trainable classifier, and GA, but not a deep generative model. Although \cite{krenn2020self} does not use a deep generative model, they train a DNN-based classifier during iterations, which is designed to distinguish high potential molecules for achieving a high reward.  JANUS-C aimed to maximize the penalized logarithm of the octanol-water partition coefficient scores \cite{gomez2018automatic}. Thus, in this comparison, the objective function based on an IC$_{50}$ value was used.
	
\end{enumerate}
Table \ref{setting_table} shows the components and test environments of each method, such as the number of iterations and valid samplings per iteration. The details on each model's parameters are in Table S2. Note that all comparison methods, except for JANUS-C, used the same reward function as that of FasterGTS. JANUS-C classifies molecules based on the reward to train its trainable classifier. Thus, for JANUS-C, molecules should have different reward values. For molecules unsatisfying the reward condition in Eq. (\ref{reward_policy}), ${exp(\alpha(-\Theta_z+y_z))}$ was used instead of one. The value is always lower than one.

\begin{table}
\caption{Components and test environments of comparing methods \label{setting_table}} 
\centering
{\begin{tabular}{@{}ccccc@{}}\toprule 
\textbf{Method } & \textbf{Components} &  \textbf{\# iteration} & \textbf{\# samplings${^\star}$}\\\midrule
RNNs      & RL, RNNs     & -            & -   \\
GPT        & RL, GPT     & -            & -   \\
MCTS-RNNs      & MCTS, RNNs    & 800         & 60 \\
ChemTS         & MCTS, RNNs   & 2000         & 25  \\
GEGL           & GA, ST, LSTM   & 50, 100      & 240 \\
JANUS-C        & GA, DNNs     & 50, 100       & 240 \\
FasterGTS-RNNs & GA, MCTS, ST, RNNs  & 100*          & 100 \\
FasterGTS-GPT    & GA, MCTS, ST, GPT & 100*          & 100 \\\midrule
\end{tabular}}
{\\\raggedright
\qquad \qquad \qquad \quad ${^\star}$ represents the average number of valid samplings per one iteration.\\
\qquad \qquad \qquad \quad * represents the average value.\\
} 
\end{table}

\section{Results}

\subsection{Experimental design}
First, the 561  cell lines used to train the  value function  ${v_{\theta}}$, and then 520 cell lines having the average absolute error less than one on the ${v_{\theta}}$ were selected for accurate evaluation of ${v_{\theta}}$. To obtain the $Z$-score of a target cell line, we need adversary cell lines to build the  IC${_{50}}$ distribution. In addition, we need an additional population to evaluate whether the target cell line's $Z$-score used in the reward is also similar to an unknown other population. Therefore, out of the 520 cell lines, 289 cell lines were randomly selected to obtain $Z$-scores during training and 232 cell lines were left to obtain an independent distribution in the evaluation (Table S3). The former $Z$-scores and the latter $Z$-scores were named `$Z$-score during training' and `$Z$-score for verification', respectively. 

For evaluation, twenty one triple-negative breast cancer (TNBC) cell lines \cite{chavez2010triple} were used for target cell lines (Table S1). Among the cancer types, TNBC is notorious for its clinical difficulty \cite{choi2012clinicopathologic}. Because of the genetic variance of different subsets of TNBC, drug treatment based on genetic profiles is an essential therapeutic strategy for TNBC \cite{bareche2018unravelling, wu2020unraveling}. 

If a generative model is highly optimized toward a target property, a generative model is likely to generate unrealistic molecules \cite{renz2020failure}. Thus, the chemical properties of the generated molecules need to be evaluated together. Thus, we further evaluated similarities between molecules generated by FasterGTS and anticancer drugs in GDSC in three properties:  the quantitative estimate of drug-likeness (QED) \cite{bickerton2012quantifying},  synthetic complexity score (SCScore) \cite{coley2018scscore}, and Fr{\'e}chet ChemNet Distance (FCD) \cite{preuer2018frechet}. The ranges of the QED and SCScore were 0 to 1 (0 = worst, 1 = best) and 1 to 5 (1 = easy to synthesize, 5 = difficult to synthesize), respectively. The FCD was used to evaluate the differences in the distribution of molecules in a generative model from the distribution of reference molecules. A high FCD value shows that the distribution of the target generative model is different from the reference distribution. In our experiment, we set 3,000 molecules from ChEMBL as the reference molecules for the FCD.

Our goal is to generate cancer sample-specific cancer drugs with a small number of samplings. Therefore, we compared each model's performance with the limited number of samplings. GA-based methods, GEGL and JANUS-C \cite{ahn2020guiding, nigam2021janus}, relied on a large number of samplings. Thus, when we measured performances, we changed the number of iterations and samplings by considering the features of each method (Table 1). For example, because JANUS-C \cite{nigam2021janus} is only based on GA, the numbers of mutation and crossover were maintained as those in their implementation, which would be a time burden to the fast search. For GEGL \cite{ahn2020guiding}, the number of samplings was set as twice of FasterGTS's to maximize their searching property, which relies on the large samplings, and the number of training iterations was equal to or more than FasterGTS's. 

For FasterGTS, the average valid samplings per iteration are 100, and iteration stopped if the number of total valid samplings was over 10,000. Therefore, the average number of iterations for FasterGTS is approximately 100 (Table \ref{setting_table}). For JANUS-C and GEGL, the number of valid samplings per iteration is approximately 240, and the number of iterations is 50 and 100. Therefore, the total number of valid samplings is approximately 12,000 and 24,000. The criterion of FasterGTS is rigid compared to those of other methods because the number of samplings from FasterGTS was always less than those of the other methods. MCTS-RNNs and ChemTS have a similar environment, such as a generative model and the number of rollout and samplings during searching. However, MCTS-RNNs re-evaluates and re-expands nodes in the fixed probability. For this reason, even though the total number of valid samplings of MCTS-RNNs and ChemTS is approximately 50,000, the total iterations of MCTS-RNNs and ChemTS are 800 and 2,000, respectively.
 
For RL based models, RNNs and GPT, we selected the top 10 molecules for every 5,000 newly generated after training for each target cell line, based on the reward (Eq. (\ref{reward_fun})). For other methods, the top 10 molecules are selected among all molecules generated during the iterations. Therefore, for each method, the number of generated molecules for 21 TNBC cell lines was 210 (Table S4). A total of 223 molecules randomly selected from ChEMBL and 223 GDSC drugs were used as references for general and cancer-specific drug properties, respectively.

\begin{table}
\centering
\caption{Reward and chemical properties of comparing methods\label{results_table}} {\begin{tabular}{@{}lccccccc@{}}\toprule 
    \multicolumn{1}{c}{\multirow{1}{*}{\textbf{Method /}}} & 
    \multicolumn{1}{c}{\multirow{2}{*}{\textbf{IC${_{50}}$}}} & 
    \multicolumn{1}{c}{\multirow{1}{*}{\textbf{$Z$-score}}} & 
    \multicolumn{1}{c}{\multirow{2}{*}{Reward}} & 
    \multicolumn{1}{c}{\multirow{1}{*}{\textbf{$Z$-score}}} & 
    \multicolumn{1}{c}{\multirow{2}{*}{\textbf{QED}}} & 
    \multicolumn{1}{c}{\multirow{2}{*}{\textbf{SCScore}}} & 
    \multicolumn{1}{c}{\multirow{2}{*}{\textbf{FCD}}}\\
    
    \multicolumn{1}{c}{\multirow{1}{*}{\textbf{Reference}}}& 
    \multicolumn{1}{c}{\multirow{1}{*}{\textbf{}}}& 
    \multicolumn{1}{c}{\multirow{1}{*}{\textbf{training}}}& 
    \multicolumn{1}{c}{\multirow{1}{*}{\textbf{}}}&
    \multicolumn{1}{c}{\multirow{1}{*}{\textbf{verification}}}& 
    \multicolumn{1}{c}{\multirow{1}{*}{\textbf{}}}& 
    \multicolumn{1}{c}{\multirow{1}{*}{\textbf{}}}&
    \multicolumn{1}{c}{\multirow{1}{*}{\textbf{}}}
   \\\midrule
   
RNNs                   & -3.45 & -0.98    & 4.86   & -1.01        & 0.28 & 4.13     & 30.85 \\
GPT                    & -2.57 & -1.42    & 5.80   & -1.53        & 0.31 & 4.38     & 32.43 \\
MCTS-RNN       & -2.43 & -1.69    & 6.05   & -1.76        & 0.46 & 3.86     & 23.63 \\
ChemTS        & -1.79 & -1.59    & 5.48   & -1.67        & 0.54 & 3.84     & 12.70 \\
GEGL-100${^\star}$      & -1.96 & -1.61    & 5.64   & -1.70        & 0.44 & 3.88     & 15.75 \\
GEGL-50${^\circ}$      & -1.82 & -1.17    & 4.57   & -1.26        & 0.46 & 3.80     & 15.21 \\
JANUS-C-100${^\star}$  & -1.19 & -2.55    & 10.43  & -2.71        & 0.16 & 3.35     & 37.00 \\
JANUS-C-50${^\circ}$  & -1.06 & -2.29    & 8.64   & -2.43        & 0.18 & 3.26     & 35.40 \\
FasterGTS-RNNs         & -2.16 & -1.70    & 6.52   & -1.78        & 0.39 & 4.19     & 18.83 \\
FasterGTS-GPT          & -2.28 & -1.75    & 6.17   & -1.85        & 0.45 & 4.21     & 18.31 \\\midrule
WO-GA${\dagger}$   & -2.25 & -1.66    & 5.90   & -1.73        & 0.37 & 4.20     & 27.24 \\
WO-ST${\ddagger}$   & -1.90 & -1.72    & 5.97   & -1.84        & 0.44 & 4.18     & 16.70 \\
WO-GA\&ST${\dagger\ddagger}$ & -2.24 & -1.24    & 4.94   & -1.32        & 0.43 & 4.14     & 21.00   \\ \midrule
ChEMBL* & - & - & - & - & 0.52 & 3.72 & 10.29 \\
GDSC* & - & - & - & - & 0.46 & 4.12 & 14.54 \\ \bottomrule
\end{tabular}}
{\\\raggedright
\qquad \qquad \, \; The scores, except FCD, are the average values of 210 molecules.\\
\qquad \qquad \, \; ${^\circ}$ means results after 50 iterations. Thus, the number of valid samplings is about 12,000.\\
\qquad \qquad \, \; ${^\star}$ means results after 100 iterations. Thus, the number of valid samplings is about 24,000.\\
\qquad \qquad \, \; ${\dagger}$ means FasterGTS-GPT without genetic algorithm.\\
\qquad \qquad \, \; ${\ddagger}$ means FasterGTS-GPT without self-train.\\
\qquad \qquad \, \; * indicates the reference drug dataset, which does not have scores related to the reward.\\}
\end{table}

\begin{figure}[]
\centering
\includegraphics[width=8cm,height=4cm]{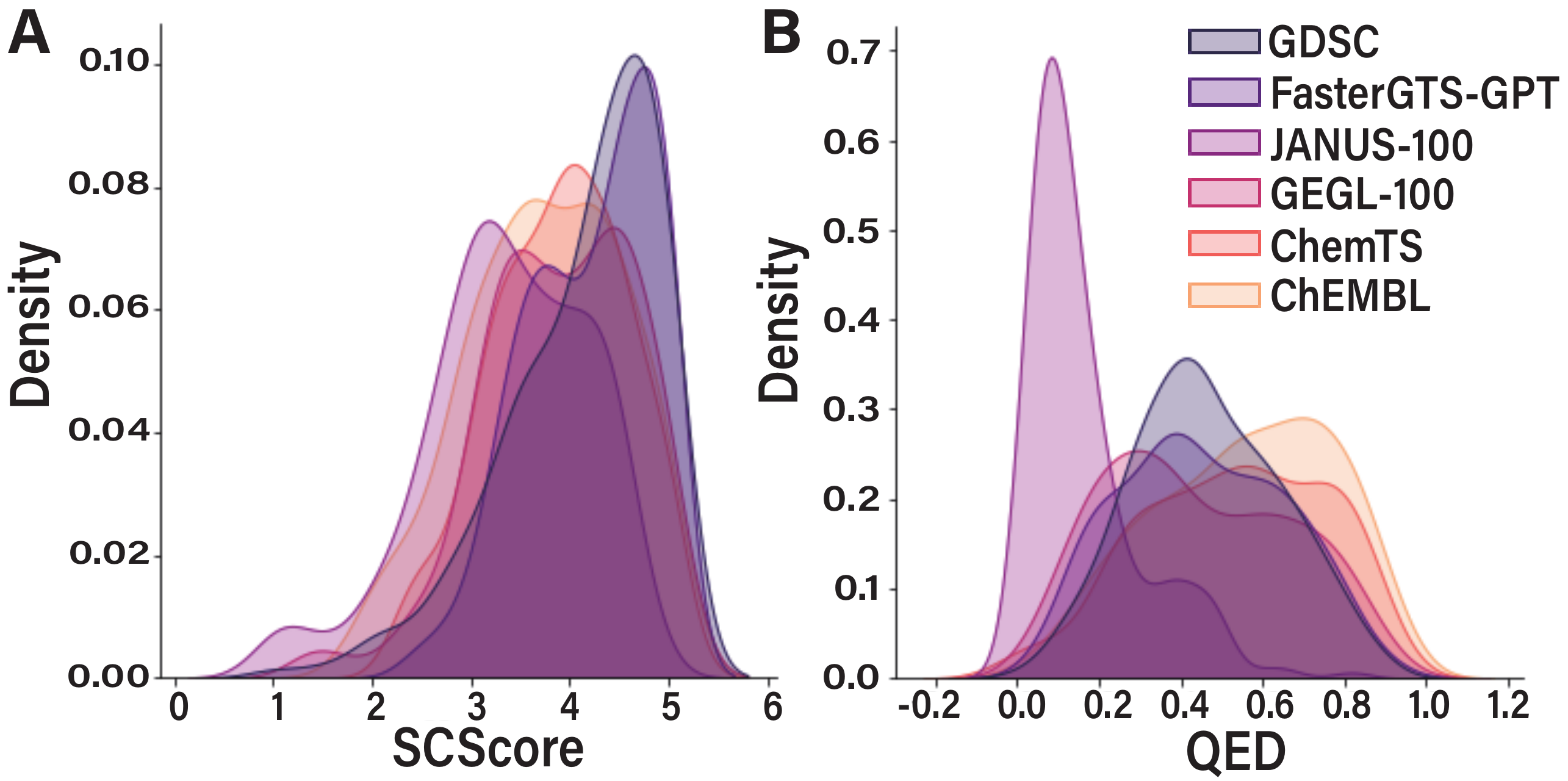} 
\caption{Distributions of methods and references for SCScore and QED. (A) SCScore distributions, where the distribution of FasterGTS-GPT is most similar to GDSC's than others, and for ChEMBL, ChemTS is. (B) QED distributions, where FasterGTS-GPT and ChemTS have the most similar distribution of reference drugs, GDSC and ChEMBL, respectively.}\label{fig3}
\end{figure}

\subsection{Reward and chemical properties}
Table 2 shows the performance of comparative methods. Performances of the methods were measured by predicted IC$_{50}$ values of generated molecules, $Z$-scores during training, rewards, $Z$-scores for verification, and three chemical properties (QED, SCScore, and FCD). Fig. 3 shows the distribution of SCScore and QED of them. To assess their chemical properties, QED, SCScore, and FCD of ChEMBL and GSDC were used as baselines.

The RL based models, RNNs and GPT, did not achieve good scores on reward and chemical properties. Although RNNs achieved the lowest IC${_{50}}$ value, it had the highest $Z$-score, meaning that the generated molecules are not specific for the given target sample. GPT showed a higher  $Z$-score than RNNs.

The MCTS-based models, ChemTS and MCTS-RNNs, showed different results, although they used the same generative model. MCTS-RNNs showed lower IC${_{50}}$ and $Z$-score values than ChemTS. In contrast, ChemTS showed better chemical property scores than MCTS-RNNs. However, the SCScore and QED scores of ChemTS are similar to those of ChEMBL, not GDSC (Figure \ref{fig3}). This result means that the model is not optimized for the target property of the target cancer sample, and the molecules from ChemTS are not different from ${p_\gamma}$ trained on ChEMBL. Even though the FCD score of MCTS-RNNs is comparatively high, MCTS-RNNs had more similar scores as those of GDSC than ChemTS for chemical properties. That means that MCTS-RNNs is not underfitted like ChemTS.

In terms of the reward, JANUS-C obtained the best score, but the model has the lowest QED and highest FCD scores. This implies that JANUS-C was highly overfitted on this task and is unlikely to generate realistic molecules to be anticancer drugs. Even though JANUS-C achieved the lowest SCScore, the score also was far from the reference scores (Figure \ref{fig3}-A). The result indicates the model generates easily synthesized but unrealistic molecules. In addition, the highest FCD score of JANUS-C represents the molecules from JANUS-Care significantly different reference drugs. For GEGL, it had good chemical properties, but low reward scores  and verified $Z$-scores than FasterGTS.

Except for JANUS, FasterGTS-GPT and -RNNs achieved the best reward score, and FasterGTS-GPT had the most similar distributions of QED and SCScore as those of GDSC (Figures \ref{fig3}, S1). Even though the FCD score of the GEGL was closer to that of GDSC than those of FasterGTS-GPT and -RNNS, FasterGTS-GPT and -RNNS had lower FCD scores than other methods. Therefore, FasterGTS-GPT could generate the most fitted anticancer drugs on a target sample, satisfying other chemical properties.

\begin{figure}[]
\centering
\includegraphics[width=8cm,height=10cm]{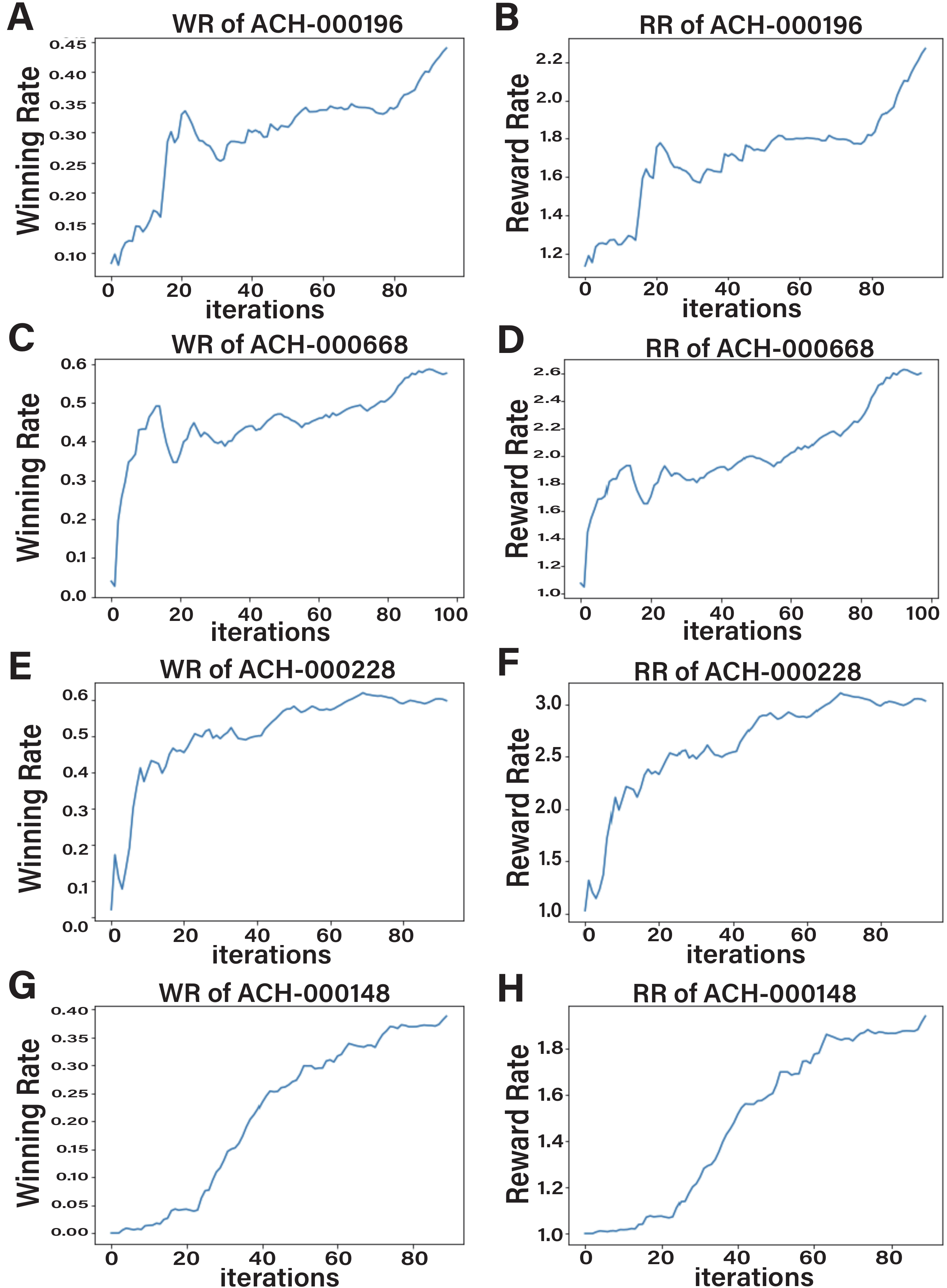}
\caption{WR and RR plots of FasterGTS-GPT for target cell lines, where x- and y-axis represent iterations and WR or RR, respectively. (A), (C), (E), and (G) are WR plots for ACH-000196, ACH-000668, ACH-000228, and ACH-000148, respectively. (B), (D), (F), and (H) are RR plots for ACH-000196, ACH-000668, ACH-000228, and ACH-000148, respectively.}\label{fig4}
\end{figure}

In general, rewards are given to an agent succeeding in a difficult task; therefore, if the task is easy, the reward becomes meaningless. The threshold values, ${\Theta_t}$ and ${\Theta_z}$, to decide whether to give a reward can be used to adjust the task level of difficulty. Herein, the WR, the number of molecules satisfying reward condition, Equation (\ref{reward_policy}), divided by the number of valid molecules, and reward rate (RR), the sum of rewards divided by the number of valid molecules, indicate the difficulty of the task and the efficiency of the agent, respectively. The WR and RR of the GD policy network ${p_\gamma}$ were  between 0.01 and 0.05, and between 1.01 and 1.09, respectively. Thus, for each target cell line, the threshold values for ${\Theta_t}$ and ${\Theta_Z}$ were set to have that the initial WR and RR of the generative network before training are between 0.01 and 0.05, and between 1.01 and 1.09, respectively (Table S1). Figures \ref{fig4} and S2,3 show how fast FasterGTS-GPT generates molecules satisfying the condition, Equation (\ref{reward_policy}), compared to the initial WR and RR in Table S1.

\subsection{Ablation study}
We conducted the ablation study to investigate how much each component of our algorithm contributes to the performance of FasterGTS-GPT. WO-GA and WO-ST achieved higher rewards than WO-GA\&ST, but lower rewards than FasterGTS-GPT. WO-ST showed a lower ${z}$-score than WO-GA and WO-GA\&ST due to using GA. In terms of drug-likeness, WO-ST was better than WO-GA and WO-GA\&ST due to using GA. These results indicate that GA increases uniqueness (low ${z}$-score) and drug-likeness (high QED and low FCD score), and ST helps to increases the reward. Taken together, we verified that it is helpful to use GA and ST for the overall performance, but their roles might be different. 

\section{Discussion}
Within the limited samplings, FasterGTS-GPT generated proper molecules in a smaller number of iterations than other methods and achieved the best performance on overall properties. Even though JANUS-C showed the highest reward scores, the method requires a large number of mutation and crossover, and their chemical properties were significantly different from those of references. In addition, more iterations progress, more serious this tendency is, when comparing JANUS-C-50 with -100. In contrast, it looks that GEGL would achieve a better result after enough iterations when comparing GEGL-50 to -100. However, GEGL-100 showed lower performance even after samplings twice more than FasterGTS-GPT. 

From the results of RNNs and GPT, the conventional RL is unlikely to generate molecules with high rewards and chemical properties. Unlike RNNs, GPT is less overfitted, which means the attention mechanism helps to alleviate the causes of overfitting, such as the vanish gradient problem. 

ChemTS achieved the highest QED score. However, it indicates that ChemTS may be underfitted for this task. Figure \ref{fig3} shows the distributions of ChemTS are similar to those of CheMBL, which are used to train ${p_\gamma}$. In contrast, MCTS-RNNs had similar scores as  chemical properties pf GDSC. Although the two methods performed until they obtain approximately 50,000 valid samplings, they do not surpass FasterGTS.

\section{Conclusion}
This study showed that our proposed model FasterGTS can generate cancer sample-specific drugs with a limited number of samplings. The experiments demonstrated that molecules generated from FasterGTS can target a specific cell line with proper drug properties. In addition, $Z$-scores for verification were always lower than $Z$-scores during training, which means that their uniqueness is robust. 

GA-based generative models showed good performance in previous studies, but required  large searching time \cite{nigam2021janus,ahn2020guiding}. In our experiment, we showed that MCTS is helpful to reduce the search time, and ST and GA contribute to increasing the reward and both drug-likeness and uniqueness, respectively. In addition, we showed that a simple MCTS is not proper to resolve complex tasks like the objective of this study, and our MCTS is more advanced than MCTS used in ChemTS.



\section*{Funding}
This work was partly supported by Institute for Information and communications Technology Promotion (IITP) grant
funded by the Korea government (MSIP) [No. 2019-0-00567, Development of Intelligent SW systems for uncovering
genetic variation and developing personalized medicine for cancer patients with unknown molecular genetic mechanisms.

\bibliographystyle{unsrt}  
\bibliography{references}

\end{document}